\documentclass[11pt]{article}

\usepackage[final]{acl}

\usepackage{times}
\usepackage{latexsym}

\usepackage[T1]{fontenc}
\usepackage[utf8]{inputenc}
\usepackage{microtype}
\usepackage{inconsolata}

\usepackage{graphicx}
\usepackage{booktabs}
\usepackage{multirow}
\usepackage{amsmath}
\usepackage{float}

\title{PSK@EEUCA 2026: Fine-Tuning Large Language Models with Synthetic Data Augmentation for Multi-Class Toxicity Detection in Gaming Chat}

\author{
  Srikar Kashyap Pulipaka \\
  Independent Researcher \\
  \texttt{srikar.kashyap@gmail.com}
}

\begin{document}
\maketitle

\begin{abstract}
This paper describes our system for the EEUCA 2026 Shared Task on Understanding Toxic Behavior in Gaming Communities. The task involves classifying World of Tanks chat messages into six toxicity categories: Non-toxic, Insults/Flaming, Other Offensive, Hate/Harassment, Threats, and Extremism. We explore multiple approaches including encoder-based models, instruction-tuned LLMs with LoRA fine-tuning, hierarchical classification, one-vs-rest strategies, and various ensemble methods. Our best system combines Llama 3.1 8B with carefully calibrated 5\% synthetic data augmentation, achieving an F1-macro score of 0.6234 on the test set, placing 4th out of 35 participating teams. We provide extensive analysis of the dataset's annotation patterns and their impact on model generalization, revealing a critical ``validation trap'' phenomenon where high validation performance correlates with poor test transfer.
\end{abstract}

\section{Introduction}

Online gaming communities face significant challenges with toxic behavior, including harassment, hate speech, and threats. The EEUCA 2026 Shared Task on Understanding Toxic Behavior in Gaming Communities \cite{thapa2026toxicity} focuses on detecting and classifying toxicity in World of Tanks chat messages, aiming to promote healthier digital spaces through AI-based moderation tools.

The task presents several unique challenges:
\begin{itemize}
    \item Extreme class imbalance (81\% Non-toxic, <1\% for rare classes)
    \item Short, informal text with gaming-specific vocabulary
    \item Multilingual content requiring cross-lingual understanding
    \item Subtle distinctions between toxic categories (e.g., skill-based insults vs. identity-based hate)
\end{itemize}

Our main strategy combines instruction-tuned LLMs (Llama 3.1 8B) with parameter-efficient fine-tuning via LoRA and carefully calibrated synthetic data augmentation. We find that a narrow 5\% synthetic data ratio is optimal, with deviations in either direction significantly degrading test performance.

Our key discovery is the ``validation trap'' phenomenon: models achieving high validation F1 through conservative predictions (matching validation distribution) perform poorly on test data. This affected our larger models most severely, with 12B models showing 0.66 validation F1 but only 0.52 test F1. Our final system achieves 0.6234 F1-macro, placing 4th overall out of 35 teams.

\section{Background}

\subsection{Task Description}

The EEUCA 2026 toxicity detection task \cite{thapa2026toxicity} is part of the 9th Workshop on Event Extraction and Understanding \cite{hurriyetouglu2026eeuca}. The task requires classifying gaming chat messages into six categories based on the annotation schema from \citet{bhandari2023crisishatemm}:

\begin{enumerate}
    \item[0.] \textbf{Non-toxic}: Normal or positive communication
    \item[1.] \textbf{Insults/Flaming}: Personal attacks targeting gaming skill
    \item[2.] \textbf{Other Offensive}: Inappropriate content not directly attacking
    \item[3.] \textbf{Hate/Harassment}: Targeted abuse based on identity
    \item[4.] \textbf{Threats}: Violence or harm threats
    \item[5.] \textbf{Extremism}: Hate ideology and dehumanization
\end{enumerate}

\subsection{Dataset}

The dataset is derived from the GameTox corpus \cite{naseem2025gametox}, comprising World of Tanks chat messages. Table \ref{tab:class_dist} shows the severe class imbalance, with Non-toxic messages comprising 81\% and rare classes (Threats, Extremism) together representing less than 0.2\%.

\begin{table}[h]
\centering
\small
\begin{tabular}{lrr}
\toprule
\textbf{Class} & \textbf{Count} & \textbf{\%} \\
\midrule
0: Non-toxic & 34,797 & 81.0\% \\
1: Insults/Flaming & 5,925 & 13.8\% \\
2: Other Offensive & 1,874 & 4.4\% \\
3: Hate/Harassment & 279 & 0.6\% \\
4: Threats & 60 & 0.1\% \\
5: Extremism & 24 & 0.1\% \\
\midrule
\textbf{Total} & 42,959 & 100\% \\
\bottomrule
\end{tabular}
\caption{Training set class distribution showing severe imbalance.}
\label{tab:class_dist}
\end{table}

Our analysis revealed significant data quality patterns: 40.2\% of training messages are exact duplicates, and 13.4\% have the same text with different labels. Interestingly, training on deduplicated data hurt performance (0.44 vs 0.60 F1), suggesting duplicates provide beneficial implicit oversampling.

\subsection{Related Work}

Toxicity detection has been extensively studied using transformer-based models \cite{devlin2019bert, liu2019roberta}. Recent work has shown that instruction-tuned LLMs can achieve strong performance on classification tasks \cite{wei2022finetuned, thapa2025large}. Parameter-efficient fine-tuning methods like LoRA \cite{hu2022lora} enable adaptation of large models with limited resources.

Gaming-specific toxicity presents unique challenges due to domain vocabulary and skill-based criticism that may or may not constitute toxicity \cite{kwak2015exploring}. Hate speech detection more broadly has been studied with various approaches \cite{parihar2021hate}.

\section{System Overview}

\subsection{Model Architecture}

We experimented with multiple architectures:

\begin{itemize}
    \item \textbf{XLM-RoBERTa Large} (560M): Full fine-tuning
    \item \textbf{Gemma 2B} \cite{team2024gemma}: LoRA + 8-bit quantization
    \item \textbf{Gemma 3 12B} \cite{team2024gemma}: LoRA + 4-bit quantization
    \item \textbf{Llama 3.1 8B} \cite{llama2024herd}: LoRA + 4-bit quantization (best)
\end{itemize}

Our final system uses Llama 3.1 8B with 4-bit NF4 quantization \cite{dettmers2023qlora} and LoRA adapters (rank=16, alpha=64).

\subsection{Prompt Engineering}

Following insights that class definitions help LLMs discriminate between similar categories, we prepend structured definitions to each input:

\begin{quote}
\small
\texttt{Classify gaming chat toxicity:}\\
\texttt{0=Non-toxic: Normal/positive chat}\\
\texttt{1=Insults: Personal attacks, slurs}\\
\texttt{2=Other Offensive: Inappropriate but not direct}\\
\texttt{3=Hate/Harassment: Targeted abuse}\\
\texttt{4=Threats: Violence/harm threats}\\
\texttt{5=Extremism: Hate ideology}\\
\texttt{Message: [input text]}
\end{quote}

This ``short'' prompt style achieved optimal balance between context and avoiding truncation.

\subsection{Synthetic Data Augmentation}

We generate synthetic training data via LLM-based paraphrase augmentation, focusing on minority classes. We used a paraphrase-only strategy after preliminary direct-generation experiments produced generic messages that did not match the short, slang-heavy style of real World of Tanks chat. Each source message was rewritten with the following template:

\begin{quote}
\small
\texttt{Rewrite this World of Tanks game chat message using different words but keeping the same meaning and toxicity level.}\\
\texttt{Original: [message]}\\
\texttt{Requirements: Keep EXACT same meaning and level of toxicity; use natural gaming language, abbreviations, slang; similar length (3--20 words). Output ONLY the rewritten message.}
\end{quote}

The synthetic pool contained 10,464 filtered paraphrases, all from minority toxicity classes: 8,348 for Class 2 (Other Offensive), 1,633 for Class 3 (Hate/Harassment), 343 for Class 4 (Threats), and 140 for Class 5 (Extremism). We applied basic cleaning, invalid-label and length filtering, label-leakage regex filtering, and embedding-based deduplication within the synthetic set. Since paraphrases are intentionally close to their source messages, we did not remove paraphrases for high similarity to the original training examples. Synthetic examples were added only to the training partition after splitting real data; validation remained 100\% real.

For the final 5\% setting, we sampled 1,921 synthetic examples from this pool (1,539 Class 2, 282 Class 3, 64 Class 4, 36 Class 5), yielding an actual synthetic share of 4.998\% of the training data. The synthetic ratio proved critical:

\begin{itemize}
    \item \textbf{5\% synthetic}: Optimal, with best test transfer
    \item 2-3\%: Insufficient, poor test transfer
    \item 7-10\%: Overfitting to synthetic patterns
    \item 15\%: Substantial degradation
\end{itemize}

The narrow optimal range suggests synthetic data helps by making predictions more ``aggressive'' on minority classes, better matching test distribution.

\section{Alternative Approaches}

We explored several alternative strategies that ultimately underperformed:

\textbf{Hierarchical Classification:} Two-stage approach (binary toxic/non-toxic, then 5-class among toxic) achieved 0.67 validation F1 but only 0.47 test F1, the largest generalization gap observed.

\textbf{One-vs-Rest:} Six binary classifiers with aggressive oversampling (up to 500x) and focal loss \cite{lin2017focal}. Too conservative at 0.56 validation F1.

\textbf{Transfer Learning:} Pre-training on DOTA 2 toxicity data before fine-tuning resulted in validation trap (0.68 val $\rightarrow$ 0.55 test).

\textbf{Ensemble Methods:} Probability averaging, voting, and confidence routing generally hurt performance because our best single model dominated all classes.

\textbf{Post-hoc Calibration:} Platt scaling, isotonic regression, and temperature scaling provided no improvement.

\section{Experimental Setup}

\subsection{Training Configuration}

\begin{itemize}
    \item Model: Llama 3.1 8B
    \item Quantization: 4-bit NF4
    \item LoRA: rank=16, alpha=64, dropout=0.0
    \item Learning rate: 5e-5 (cosine schedule)
    \item Epochs: 4
    \item Batch size: 4 (gradient accumulation: 4)
    \item Loss: class-weighted cross-entropy
    \item Synthetic ratio: 5\%
    \item Max sequence length: 384
\end{itemize}

\subsection{Evaluation}

The official metric is macro-averaged F1 score across all six classes. We used the provided validation split for development and hyperparameter tuning.

\section{Results}

\subsection{Main Results}

Table \ref{tab:results} compares our approaches. Llama 3.1 8B with 5\% synthetic data achieves the best test performance.
The unboosted 5\% synthetic model scored 0.6232; a small post-hoc Class 2 boost increased the official submitted score to 0.6234.

\begin{table}[h]
\centering
\small
\begin{tabular}{lcc}
\toprule
\textbf{System} & \textbf{Val F1} & \textbf{Test F1} \\
\midrule
XLM-RoBERTa Large & 0.30 & -- \\
Gemma 2B & 0.63 & 0.52 \\
Gemma 12B & 0.66 & 0.52 \\
Two-stage & 0.67 & 0.47 \\
Llama 8B (no synth) & 0.6554 & 0.5971 \\
\textbf{Llama 8B + 5\% synth} & \textbf{0.6271} & \textbf{0.6234} \\
\bottomrule
\end{tabular}
\caption{System comparison. Best test result in bold.}
\label{tab:results}
\end{table}

\subsection{Synthetic Data Ablation}

Table \ref{tab:synth} shows the critical sensitivity to synthetic ratio.

\begin{table}[h]
\centering
\small
\begin{tabular}{lcc}
\toprule
\textbf{Synth Ratio} & \textbf{Val F1} & \textbf{Test F1} \\
\midrule
0\% & 0.6554 & 0.5971 \\
2\% & 0.6247 & 0.5042 \\
3\% & 0.6051 & 0.5514 \\
\textbf{5\%} & \textbf{0.6271} & \textbf{0.6232} \\
7\% & 0.6214 & 0.4649 \\
10\% & 0.5499 & 0.5851 \\
15\% & 0.6045 & 0.5343 \\
\bottomrule
\end{tabular}
\caption{Effect of synthetic data ratio on Llama 8B.}
\label{tab:synth}
\end{table}

To understand why 5\% transferred best, we compared test prediction distributions for the Llama 8B models in Table \ref{tab:pred_dist}. The 5\% model reduced Non-toxic predictions and increased predictions for Classes 2 and 3, the confusable minority categories most affected by the train/test annotation shift. Higher synthetic ratios did not preserve this balance in class-level decisions and reduced test F1.

\begin{table}[h]
\centering
\small
\begin{tabular}{lccc}
\toprule
\textbf{Prediction} & \textbf{0\% synth} & \textbf{5\% synth} & \textbf{10\% synth} \\
\midrule
Class 0: Non-toxic & 79.6\% & 79.0\% & 78.7\% \\
Class 1: Insults & 14.8\% & 14.3\% & 13.9\% \\
Class 2: Other & 4.9\% & 5.7\% & 6.6\% \\
Class 3: Hate & 0.6\% & 0.7\% & 0.6\% \\
\midrule
Test F1 & 0.5971 & \textbf{0.6232} & 0.5851 \\
\bottomrule
\end{tabular}
\caption{Test prediction distribution for Llama 8B synthetic-data variants.}
\label{tab:pred_dist}
\end{table}

\subsection{Per-class Performance}

Table \ref{tab:perclass} shows per-class test F1 for the final submitted system. Performance correlates roughly with class frequency, with Class 2 (Other Offensive) and Class 3 (Hate/Harassment) being particularly challenging.

\begin{table}[h]
\centering
\small
\begin{tabular}{lc}
\toprule
\textbf{Class} & \textbf{Test F1} \\
\midrule
0: Non-toxic & 0.94 \\
1: Insults/Flaming & 0.74 \\
2: Other Offensive & 0.44 \\
3: Hate/Harassment & 0.43 \\
4: Threats & 0.33 \\
5: Extremism & 0.86 \\
\bottomrule
\end{tabular}
\caption{Per-class F1 for the final submitted system.}
\label{tab:perclass}
\end{table}

\section{Analysis}

\subsection{The Validation Trap}

Our most significant finding is the ``validation trap'': models achieving high validation F1 through conservative predictions (matching the 81\% Non-toxic distribution) performed poorly on test. Evidence includes:

\begin{itemize}
    \item Gemma 12B: 0.66 val $\rightarrow$ 0.52 test
    \item Transfer learning: 0.68 val $\rightarrow$ 0.55 test
    \item Two-stage: 0.67 val $\rightarrow$ 0.47 test
\end{itemize}

Models predicting more minority classes (2, 3) performed better on test, suggesting different annotation patterns between splits.

\subsection{Why 5\% Synthetic Works}

The 5\% ratio appears to increase minority class predictions without overwhelming original patterns. The distribution analysis in Table \ref{tab:pred_dist} supports this interpretation: relative to the no-synthetic Llama 8B model, the 5\% model predicts fewer Non-toxic messages and more Class 2/3 messages, which improves test transfer. Higher synthetic ratios did not yield the same class-level accuracy: the 10\% model shifted predictions further toward Class 2 but lost roughly 0.038 test F1, suggesting that excessive synthetic data can reinforce artifacts or shift the model away from the test annotation pattern.

\subsection{Error Analysis}

Common error patterns include:
\begin{itemize}
    \item Confusion between Class 1 (Insults) and Class 2 (Other Offensive)
    \item Multilingual messages misclassified as Non-toxic
    \item Gaming slang incorrectly flagged as toxic
\end{itemize}

\section{Conclusion}

We presented a comprehensive exploration of approaches for gaming toxicity detection. Key findings:

\begin{enumerate}
    \item Llama 3.1 8B outperformed both smaller and larger models
    \item Synthetic data has a narrow sweet spot (5\%)
    \item Validation metrics can be misleading due to distribution shift
    \item Ensembles don't help when one model dominates
\end{enumerate}

Our system achieves 0.6234 F1-macro, placing 4th out of 35 teams. Future work could explore better handling of distribution shift and external gaming-specific data.

\section*{Limitations}

Our analysis is limited to this specific dataset. The ``validation trap'' phenomenon may be dataset-specific and not generalize. Computational constraints limited exploration of larger models and longer training. The synthetic data approach requires access to commercial LLM APIs.

\section*{Ethics Statement}

This work involves detecting toxic content in gaming chat. Models could potentially be misused to generate toxic content or for surveillance. We advocate for responsible deployment in content moderation systems with human oversight, transparency about automated decisions, and appeal mechanisms for users.

\bibliography{references}

\clearpage
\appendix

\section{Full Test Performance}

Table \ref{tab:full_test_report} reports the full test-set classification report for the final submitted system. These scores were computed after the official test labels were released, using the submitted predictions that achieved 0.6234 macro-F1.

\begin{table}[H]
\centering
\small
\begin{tabular}{lrrrr}
\toprule
\textbf{Class} & \textbf{Precision} & \textbf{Recall} & \textbf{F1} & \textbf{Support} \\
\midrule
0: Non-toxic & 0.9620 & 0.9242 & 0.9427 & 4351 \\
1: Insults/Flaming & 0.7563 & 0.7318 & 0.7438 & 742 \\
2: Other Offensive & 0.3396 & 0.6128 & 0.4370 & 235 \\
3: Hate/Harassment & 0.4103 & 0.4444 & 0.4267 & 36 \\
4: Threats & 0.3000 & 0.3750 & 0.3333 & 8 \\
5: Extremism & 0.7500 & 1.0000 & 0.8571 & 3 \\
\midrule
Macro average & 0.5864 & 0.6814 & 0.6234 & 5375 \\
Weighted average & 0.9016 & 0.8800 & 0.8887 & 5375 \\
\bottomrule
\end{tabular}
\caption{Full test-set classification report for the final submitted system.}
\label{tab:full_test_report}
\end{table}

\section{Additional Experimental Results}

Table \ref{tab:appendix_experiments} summarizes additional systems and ablations explored during development. The pattern reinforces the main paper's validation-trap finding: several systems improved validation F1 but transferred poorly to the test set, while the final Llama 8B system with a small amount of synthetic data gave the best test performance.

\begin{table}[H]
\centering
\scriptsize
\begin{tabular}{p{0.34\linewidth}ccp{0.34\linewidth}}
\toprule
\textbf{System} & \textbf{Val F1} & \textbf{Test F1} & \textbf{Notes} \\
\midrule
Zero-shot GPT-4o-mini & 0.4630 & 0.4126 & Direct prompting; over-predicted minority classes \\
Two-stage Gemma 2B & 0.6749 & $\sim$0.47 & Binary toxic detector plus toxic-only classifier \\
Gemma 2B & 0.63 & $\sim$0.52 & Single-stage LoRA baseline \\
Gemma 12B & 0.662 & $\sim$0.52 & Higher validation F1 but conservative test predictions \\
Prompted ensemble & 0.6201 & 0.5762 & Average of prompted 2B models \\
Multi-step ensemble & 0.6280 & 0.5810 & Confidence-based routing \\
Gemma 2B train-all & -- & 0.5898 & Trained on combined train and validation data \\
Llama 8B, no synthetic & 0.6554 & 0.5971 & Best single model before augmentation \\
Llama 8B + 10\% synthetic & $\sim$0.65 & 0.5851 & Higher synthetic ratio hurt transfer \\
Transfer DOTA2 $\rightarrow$ GameTox & 0.6815 & $\sim$0.55 & Gaming-domain pretraining caused validation trap \\
Llama 8B + 5\% synthetic & 0.6271 & 0.6232 & Best unboosted model \\
Final Class 2 boost & -- & \textbf{0.6234} & Official submitted system \\
\bottomrule
\end{tabular}
\caption{Additional systems and ablations evaluated during development.}
\label{tab:appendix_experiments}
\end{table}

\end{document}